%% file: eacl2017.tex
\documentclass[11pt]{article}
\input{./_metadata/preamble.txt}
\eaclfinalcopy 
\def\eaclpaperid{21} 

\title{One Representation per Word --- \emph{Does it make Sense for Composition?}}

\author{
  Thomas Kober, Julie Weeds, John Wilkie, Jeremy Reffin \and David Weir\\
 TAG laboratory, Department of Informatics, University of Sussex\\
 Brighton, BN1 9RH, UK\\
  {\tt \{t.kober, j.e.weeds, jw478, j.p.reffin, d.j.weir\}@sussex.ac.uk}
}
\date{}

\begin{document}
\maketitle

\begin{abstract}
\input{./abstract.tex}
\end{abstract}

\section{Introduction}
\label{sec:introduction}
\input{./introduction.tex}

\section{Evaluating Distributional Models of Composition}
\label{sec:models}
\input{./models.tex}

\section{Phrase Similarity}
\label{sec:evaluation}
\input{./evaluation.tex}

\section{Word Sense Discrimination}
\label{sec:task}
\label{sec:task_description}
\input{./task.tex}

\section{Discussion}
\label{sec:discussion}
\input{./discussion.tex}

\section{Related Work}
\label{sec:related_work}
\input{./relatedwork.tex}

\section{Conclusion}
\label{sec:conclusion}
\input{./conclusion.tex}

\section*{Acknowledgments}
\label{acknowledgments}
\input{./acknowledgments.tex}

\bibliography{common}
\bibliographystyle{eacl2017}

\end{document}

%% file: _metadata/preamble.txt
\usepackage{times}
\usepackage{url}
\usepackage{latexsym}
\usepackage{subcaption}
\usepackage{graphicx}
\usepackage{placeins}
\usepackage[draft]{hyperref}
\usepackage{minibox}
\usepackage{eacl2017}
\usepackage{footmisc}

\DeclareCaptionFont{tiny}{\tiny}

\usepackage{amsmath,amsthm,amssymb,amsfonts,mathtools}

\usepackage{tabularx}
\usepackage{multirow}
\usepackage{wrapfig}

\usepackage[justification=justified,singlelinecheck=false]{caption}

\usepackage{algorithm}
\usepackage{algpseudocode}

\usepackage{tikz-dependency}

\newcommand{\Sensembed}{%
	\textsc{Sens}\textsc{Embed}}


%% file: abstract.tex
In this paper, we investigate whether an \emph{a priori} disambiguation of word senses is strictly necessary or whether the meaning of a word in context can be disambiguated through composition alone. We evaluate the performance of off-the-shelf single-vector and multi-sense vector models on a benchmark phrase similarity task and a novel task for word-sense discrimination. We find that single-sense vector models perform as well or better than multi-sense vector models despite arguably less clean elementary representations. Our findings furthermore show that simple composition functions such as pointwise addition are able to recover sense specific information from a single-sense vector model remarkably well.

%% file: introduction.tex
Distributional word representations based on counting co-occurrences have a long history in natural language processing and have successfully been applied to numerous tasks such as sentiment analysis, recognising textual entailment, word-sense disambiguation and many other important problems. More recently low-dimensional and dense neural word embeddings have received a considerable amount of attention in the research community and have become ubiquitous in numerous NLP pipelines in academia and industry. One fundamental simplifying assumption commonly made in distributional semantic models, however, is that every word can be encoded by a single representation. Combining polysemous lexemes into a single vector has the consequence of essentially creating a weighted average of all observed meanings of a lexeme in a given text corpus.

Therefore a number of proposals have been made to overcome the issue of conflating several different senses of an individual lexeme into a single representation.  One approach~\cite{Reisinger_2010,Huang_2012} is to try directly inferring a predefined number of senses from data and subsequently label any occurrences of a polysemous lexeme with the inferred inventory. Similar approaches are proposed by Reddy et al.~\shortcite{Reddy_2011b} and Kartsaklis et al.~\shortcite{Kartsaklis_2013a} who show that appropriate sense selection or disambiguation typically improves performance for composition of noun phrases~\cite{Reddy_2011b} and verb phrases~\cite{Kartsaklis_2013a}. Dinu and Lapata~\shortcite{Dinu_2010} proposed a model that represents the meaning of a word as a probability distribution over latent senses which is modulated based on contextualisation, and report improved performance on a word similarity task and the lexical substitution task. Other approaches leverage an existing lexical resource such as BabelNet or WordNet to obtain sense labels \emph{a priori} to creating word representations~\cite{Iacobacci_2015}, or as a postprocessing step after obtaining initial word representations~\cite{Chen_2014,Pilehvar_2016}. While these approaches have exhibited strong performance on benchmark word similarity tasks~\cite{Huang_2012,Iacobacci_2015} and some downstream processing tasks such as part-of-speech tagging and relation identification~\cite{Li_2015}, they have been weaker than the single-vector representations at predicting the compositionality of multi-word expressions~\cite{Salehi_2015}, and at tasks which require the meaning of a word to be considered in context; e.g, word sense disambiguation~\cite{Iacobacci_2016} and word similarity in context~\cite{Iacobacci_2015}. 

In this paper we consider what happens when distributional representations are composed to form representations for larger units of meaning. In a compositional phrase, the meaning of the whole can be inferred from the meaning of its parts.  Thus, assuming compositionality, the representation of a phrase such as \emph{black mood}, should be directly inferable from the representations for \emph{black} and for \emph{mood}.  Further, one might suppose that composing the correct senses of the individual lexemes would result in a more accurate representation of that phrase.  However, our counter-hypothesis is that the act of composition contextualises or disambiguates each of the lexemes thereby making the representations of individual senses redundant.   We investigate this hypothesis by evaluating the performance of single-vector representations and multi-sense representations at both a benchmark phrase similarity task and at a novel word-sense discrimination task.

Our contributions in this work are thus as follows.  First, we provide quantitative and qualitative evidence that even simple composition functions have the ability to recover sense-specific information from a single-vector representation of a polysemous lexeme in context.  Second, we introduce a novel word-sense discrimination task\footnote{Our task is available from \url{https://github.com/tttthomasssss/sense2017}}, which can be seen as the first stage of word-sense disambiguation. The goal is to find whether the occurrences of a lexeme in two or more sentential contexts belong to the same sense or not, without necessarily labelling the senses. While it has received relatively little attention in recent years, it is an important natural language understanding problem and can provide important insights into the process of semantic composition.


%% file: models.tex
For evaluation we use several readily available off-the-shelf word embeddings, that have already been shown to work well for a number of different NLP applications. We compare the 300-dimensional skip-gram \texttt{word2vec}~\cite{Mikolov_2013b} word embeddings\footnote{Available from: \url{https://code.google.com/p/word2vec/}} to the dependency based version of \texttt{word2vec} --- henceforth \texttt{dep2vec}\footnote{Available from: \url{https://levyomer.wordpress.com/2014/04/25/dependency-based-word-embeddings/}}~\cite{Levy_2014} --- and the {\textsc{Sens}\textsc{Embed}} model\footnote{Available from: \url{http://lcl.uniroma1.it/sensembed/}} by Iacobacci et al.~\shortcite{Iacobacci_2015}, which creates word-sense embeddings by performing word-sense disambiguation prior to running \texttt{word2vec}. 


We note that \texttt{word2vec} and \texttt{dep2vec} use a single vector per word approach and therefore conflate the different senses of a polysemous lexeme.  On the other hand, {\textsc{Sens}\textsc{Embed}} utilises Babelfy~\cite{Moro_2014} as an external knowledge source to perform word-sense disambiguation and subsequently creates one vector representation per word sense. 

For composition we use pointwise addition for all models as this has been shown to be a strong baseline in a number of studies~\cite{Hashimoto_2014,Hill_2016}. We also experimented with pointwise multiplication as composition function but, similar to Hill et al.~\shortcite{Hill_2016}, found its performance to be very poor (results not reported). We model any out-of-vocabulary items as a vector consisting of all zeros and determine proximity of composed meaning representations in terms of cosine similarity. We lowercase and lemmatise the data in our task but do not perform number or date normalisation, or removal of rare words. 


%% file: evaluation.tex
Our first evaluation task is the benchmark phrase similarity task of Mitchell and Lapata~\shortcite{Mitchell_2010}.  This dataset consists of 108 adjective-noun (AN), 108 noun-noun (NN) and 108 verb-object (VO) pairs.  The task is to compare a compositional model's similarity estimates with human judgements by computing Spearman's $\rho$.  An average $\rho$ of 0.47-0.48 represents the current state-of-the-art performance on this task \cite{Hashimoto_2014,Kober_2016,Wieting_2015}.

For single-sense representations, the strategy for carrying out this task is simple.  For each phrase in each pair, we compose the constituent representations and then compute the similarity of each pair of phrases using the cosine similarity.  For multi-sense representations, we adapted the strategy which has been used successfully in various word similarity experiments \cite{Huang_2012,Iacobacci_2015}.  Typically, for each word pair, all pairs of senses are considered and the similarity of the word pair is considered to be the similarity of the closest pair of senses.  The fact that this strategy works well suggests that when humans are asked to judge word similarity, the pairing automatically primes them to select the closest senses.  Extending this to phrase similarity requires us to compose each possible pair of senses for each phrase and then select the sense configuration which results in maximal phrase similarity.  For comparison, we also give results for the configuration which results in minimal phrase similarity and the arithmetic mean\footnote{We also tried the geometric mean and the median but these performed comparably with the arithmetic mean.}
 of all sense configurations.  
\subsection{Results}

\begin{table}[ht]
\small
\begin{tabular}{l|c|c|c|c}
Model&AN&NN&VO&Average\\
\hline
\textbf{\texttt{word2vec}}& 0.47 & \textbf{0.46} & \textbf{0.45} & \textbf{0.46} \\
\textbf{\texttt{dep2vec}}& \textbf{0.48} & \textbf{0.46} & \textbf{0.45} & \textbf{0.46} \\
\textbf{{\textsc{Sens}\textsc{Embed}}}:max & 0.39 & 0.39 & 0.32 & 0.37 \\
\textbf{{\textsc{Sens}\textsc{Embed}}}:min & 0.24 & 0.12 & 0.22 & 0.19 \\
\textbf{{\textsc{Sens}\textsc{Embed}}}:mean & 0.46 & 0.35 & 0.37 & 0.39 \\
\end{tabular}
\captionsetup{singlelinecheck=false,font=small,labelsep=newline}
\caption{Results for the Mitchell and Laptata~\shortcite{Mitchell_2010} dataset.}
\label{tab:ml10}
\end{table}

Table~\ref{tab:ml10} shows that the simple strategy of adding high quality single-vector representations is very competitive with the state-of-the-art for this task.  None of the strategies for selecting a sense configuration for the multi-sense representations could compete with the single sense representations on this task.  One possible explanation is that the commonly adopted closest sense strategy is not effective for composition since the composition of incorrect senses may lead to spuriously high similarities (for two ``implausible" sense configurations). 

Table~\ref{high_sim_examples} lists a number of example phrase pairs with low average human similarity scores in the Mitchell and Lapata~\shortcite{Mitchell_2010} test set. The results show the tendency of the closest sense strategy with \Sensembed~(SE) to overestimate the similarity of dissimilar phrase pairs. For a comparison we manually labelled the lexemes in the sample phrases with the appropriate BabelNet senses prior to composition (SE*). Human (H) similarity scores are normalised and averaged for an easier comparison, model estimates represent cosine similarities. 
\begin{table}[ht]
\small
\resizebox{\columnwidth}{!}{
\begin{tabular}{l | l | c | c | c}
\textbf{Phrase 1} & \textbf{Phrase 2} & \textsc{\textbf{SE}} & \textbf{SE*} & \textbf{H}\\
\hline
\emph{buy land} & \emph{leave house} & 0.49 & 0.28 & 0.26 \\
\emph{close eye} & \emph{stretch arm} & 0.40 & 0.31 & 0.25 \\
\emph{wave hand} & \emph{leave company} & 0.42 & 0.08 & 0.20 \\
\emph{drink water} & \emph{use test} & 0.29 & 0.04 & 0.18 \\
\emph{european state} & \emph{present position} & 0.28 & -0.03 & 0.19 \\
\emph{high point} & \emph{particular case} & 0.41 & 0.10 & 0.21
\end{tabular}}
\captionsetup{singlelinecheck=false,font=small,labelsep=newline}
\caption{Tendency of \Sensembed~(SE) to overestimate the similarity on phrase pairs with low average human similarity when the closest sense strategy is used.}
\label{high_sim_examples}
\end{table}

%% file: task.tex
Word-sense discrimination can be seen as the first stage of word-sense disambiguation, where the goal is to find whether two or more occurrences of the same lexeme express identical senses, without necessarily labelling the senses yet. It has received relatively little attention despite its potential for providing important insights into semantic composition, focusing in particular on to the ability of compositional distributional semantic models to appropriately contextualise a polysemous lexeme.

Work on word-sense discrimination has suffered from the absence of a benchmark task as well as a clear evaluation methodology. For example Sch\"{u}tze~\shortcite{Schutze_1998} evaluated his model on a dataset consisting of 20 polysemous words (10 naturally ambiguous lexemes and 10 artificially ambiguous ``pseudo-lexemes") in terms of accuracy for coarse grained sense distinctions, and an information retrieval task. Pantel and Lin~\shortcite{Pantel_2002}, and Van de Cruys~\shortcite{VanDeCruys_2008} used automatically extracted words from various newswire sources and evaluated the output of their models in comparison to WordNet and EuroWordNet, respectively. Purandare and Pedersen~\shortcite{Purandare_2004} used a subset of the words from the \textsc{Senseval-2} task and evaluated their models in terms of precision, recall and F1-score of how well available sense tags match with clusters discovered by their algorithms. Akkaya et al.~\shortcite{Akkaya_2012} used the concatenation of the \textsc{Senseval-2} and \textsc{Senseval-3} tasks and evaluated their models in terms of cluster purity and accuracy. Finally, Moen et al.~\shortcite{Moen_2013} used the semantic textual similarity (STS) 2012 task, which is based on human judgements of the similarity between two sentences. 

One contribution of our work is a novel word-sense discrimination task, evaluated on a number of robust baselines in order to facilitate future research in that area. In particular, our task offers a testbed for assessing the contextualisation ability of compositional distributional semantic models. The goal is, for a given polysemous lexeme in context, to identify the sentence from a list of options that is expressing the same sense of that lexeme as the given target sentence. These two sentences --- the target and the ``correct answer" --- can exhibit any degree of semantic similarity as long as they convey the same sense of the target lexeme. 
Table~\ref{task_example} shows an example of the polysemous adjective \emph{black} in our task. The goal of any model would be to determine that the expressed sense of \emph{black} in the sentence \emph{She was going to set him free from all of the evil and black hatred he had brought to the world} is identical to the expressed sense of \emph{black} in the target sentence \emph{Or should they rebut the Democrats' black smear campaign with the evidence at hand}.
\begin{table*}[!htb]
\centering
\small
\resizebox{\textwidth}{!}{
\begin{tabular}{l | l | l }
& \textbf{Sense Definition} & \textbf{Sentence} \\ \hline
\textbf{Target} & full of anger or hatred & Or should they rebut the Democrats' \textbf{black} smear campaign with \\ && the evidence at hand? \\ \hline\hline
Option 1 & full of anger or hatred & She was going to set him free from all of the evil and \textbf{black} hatred \\ && he had brought to the world. \\\hline
Option 2 & (of a person's state of mind) & I've been in a \textbf{black} mood since September 2001, it's hanging over \\ & full of gloom or misery; very depressed  & me like a penumbra.\\\hline
Option 3 & (of humour) presenting tragic or harrowing & Over the years I have come to believe that fate either hates me, or \\ & situations in comic terms & has one hell of a \textbf{black} sense of humour.\\\hline
Option 4 & (of coffee or tea) served without milk & The young man was reading a paperback novel and sipping a \\ && steaming mug of hot, \textbf{black} coffee.\\
\end{tabular}}
\captionsetup{font=small}
\caption{Example of the polysemous adjective \emph{black} in our task. The goal for any model is to predict option 1 as expressing the same sense of \emph{black} as the target sentence.}
\label{task_example}
\end{table*}

Our task assesses the ability of a model to discriminate a particular sense in a sentential context from any other senses and thus provides an excellent testbed for evaluating multi-sense word vector models as well as compositional distributional semantic models.   By composing the representation of a target lexeme with its surrounding context, it should be possible to determine its sense.  For example, composing \emph{black smear campaign} should lead to a compositional representation that is closer to the composed representation of \emph{black hatred} than to \emph{black mood}, \emph{black sense of humour} or \emph{black coffee}. This essentially uses the similarity of the compositional representation of a lexeme's context to determine its sense. Similar approaches to word-sense disambiguation have already been successfully used in past works~\cite{Akkaya_2012,Basile_2014}.

\subsection{Task Construction}
\label{subsec:dataset_construction}
For the construction of our dataset we made use of data from two english dictionaries (Oxford Dictionary and Collins Dictionary), accessible via their respective web APIs\footnote{\url{https://developer.oxforddictionaries.com} for the Oxford Dictionary, \url{https://www.collinsdictionary.com/api/} for the Collins Dictionary. We use NLTK 3.2 to access SemCor.}, as well as examples from the sense annotated corpus SemCor~\cite{Miller_1993}. Our use of dictionary data is motivated by a number of favourable properties which make it a very suitable data source for our proposed task:
\begin{itemize}
	\item The content is of very high-quality and curated by expert lexicographers.
	\item All example sentences are carefully crafted in order to unambiguously illustrate the usage of a particular sense for a given polysemous lexeme.
	\item The granularity of the sense inventory reflects common language use\footnote{The Oxford dictionary lists 5 different senses for the noun ``bank", whereas WordNet 3.0 lists 10 synsets, for example distinguishing ``bank" as the concept for a financial institution and ``bank" as a reference to the building where financial transactions take place.}.
	\item The example sentences are typically free of any domain bias wherever possible.
	\item The data is easily accessible via a web API.
\end{itemize}
By using the data from curated resources we were able to avoid a setup as a sentence similarity task and any potentially noisy crowd-sourced human similarity judgements. 

\begin{figure}[!htb]
\includegraphics[width=\columnwidth]{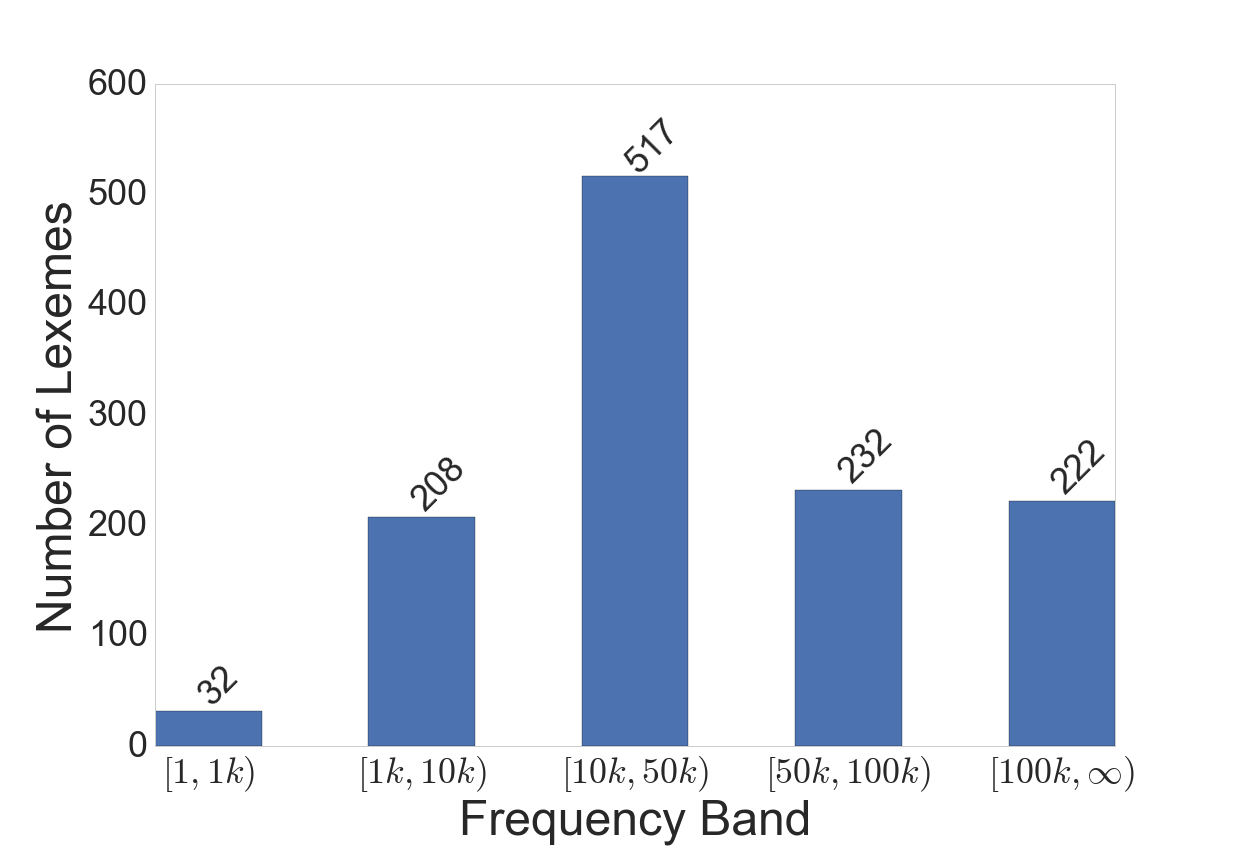}
\captionsetup{font=small}
\caption{Binned frequency distribution of the polysemous target lexemes in our task.}
\label{lexeme_frequencies}
\end{figure}

We were furthermore able to collect data from varying frequency bands, enabling an assessment of the impact of frequency on any model. Figure~\ref{lexeme_frequencies} shows the number of target lexemes per frequency band. While the majority of lexemes, with reference to a cleaned October 2013 Wikipedia dump\footnote{We removed any articles with fewer than 20 page views.}, is in the middle band, there is a considerable amount of less frequent lexemes. The most frequent target lexeme in our task is the verb \emph{be} with $\approx$$1.8$m occurrences in Wikipedia, and the least frequent lexeme is the verb \emph{ruffle} with only $57$ occurrences. The average target lexeme frequency is $\approx$$95$k for adjectives, and $\approx$$45$k$-46$k for nouns and  verbs\footnote{The overall number of unique word types is smaller than the number of examples in our task as there are a number of lexemes that can occur with more than one part-of-speech.}.

\subsection{Task Setup Details}
\label{subsec:task_setup_details}
We collected data for 3 different parts-of-speech: adjectives, nouns and verbs. We furthermore created task setups with varying numbers of senses to distinguish (2-5 senses) for a given target lexeme. This is to evaluate how well a model is able to discriminate different degrees of polysemy of any lexeme. For any task setup evaluating for $n$ senses, we included all lexemes with $>n$ senses and randomly sampled $n$ senses from its inventory. For each lexeme, we furthermore ensured that it had at least 2 example sentences per sense. For the available senses of any given lexeme, we randomly chose a sense as the target sense, and from its list of example sentences randomly sampled 2 sentences, one as the target example and one as the ``correct answer" for the list of candidate sentences. Finally we once again randomly sampled the required number of other senses and example sentences to complete the task setup. Using random sampling of word senses and targets aims to avoid a predominant sense bias.

For each part-of-speech we created a development split for parameter tuning and a test split for the final evaluation. Table~\ref{dataset_stats} shows the number of examples for each setup variant of our task. The biggest category are polysemous nouns, representing roughly half of the data, followed by verbs representing another third, and the smallest category are adjectives taking up the remaining $\approx$$17\%$.
\begin{table}[!htb]
\centering
\small
\resizebox{\columnwidth}{!}{
\begin{tabular}{l | c | c | c | c }
&\textbf{2 senses} & \textbf{3 senses} & \textbf{4 senses} & \textbf{5 senses} 	\\ \hline
\textbf{Adjective} 	& 66/209 		& 47/170 		& 37/137 		& 28/115	\\
\textbf{Noun} 		& 170/618 	& 125/499 	& 100/412 	& 74/345	\\
\textbf{Verb} 		& 127/438 	& 71/354 		& 72/295 		& 56/256	\\ \hline
\textbf{Total} 		& 363/1265	& 263/1023	& 209/844		& 164/716	\\
\end{tabular}}
\captionsetup{font=small}
\caption{Number of examples per part-of-speech and number of senses (\emph{\#dev examples}/\emph{\#train examples}). }
\label{dataset_stats}
\end{table}
We measure performance in terms of accuracy of correctly predicting which two sentences share the same sense of a given target lexeme. Accuracy has the advantage of being much easier to interpret --- in absolute terms as well as in the relative difference between models --- in comparison to other commonly used evaluation metrics such as cluster purity measures or correlation metrics such as Spearman $\rho$ and Pearson $r$.

\subsection{Experimental Setup}

In this paper we compare the compositional models outlined earlier with two baselines, a random baseline and a word-overlap baseline of the extracted contexts.  For the single-vector representations, we composed the target lexeme with all of the words in the context window and compared it with the equivalent representation of each of the options (lexeme plus context words).  The option with the highest cosine similarity was deemed to be the selected sense. For {\textsc{Sens}\textsc{Embed}}, we composed all sense vectors of a target lexeme with the given context and then used the closest sense strategy~\cite{Iacobacci_2015} on composed representations to choose the predicted sense\footnote{We also tried an all-by-all senses composition, however found this to be computationally not tractable.}.  The word-overlap baseline is simply the number of words in common between the context window for the target and each of the options.

We experimented with symmetric linear bag-of-words contexts of size 1, 2 and 4 around the target lexeme. We also experimented with dependency contexts, where first-order dependency contexts performed almost identical to using a 2-word bag-of-words context window (results not reported). We excluded stop words prior to extracting the context window in order to maximise the number of content words.  We break ties for any of the methods --- including the baselines --- by randomly picking one of the options with the highest similarity to the composed representation of the target lexeme with its context. Statistical significance between the best performing model and the word overlap baseline is computed by using a randomised pairwise permutation test~\cite{Efron_1994}.

\subsection{Results}

Table~\ref{results_linear_window_all} shows the results for all context window sizes across all parts-of-speech and number of senses. All models substantially outperform the random baseline for any number of senses. Interestingly the word overlap baseline is competitive for all context window sizes. While it is a very simple method, it has already been found to be a strong baseline for paraphrase detection and semantic textual similarity~\cite{Dinu_2012}. One possible explanation for its robust performance on our task is an occurrence of the one-sense-per-collocation hypothesis~\cite{Yarowsky_1993}. The performance of all other models is roughly in the same ballpark for all parts-of-speech and number of senses, suggesting that they form robust baselines for future models. 
\input{./_tables/bow_1-4_no_adverbs.tex}
While the results are relatively mixed for adjectives, \texttt{word2vec} appears to be the strongest model for polysemous nouns and verbs. 

\begin{figure*}[!htb]
\includegraphics[width=\textwidth]{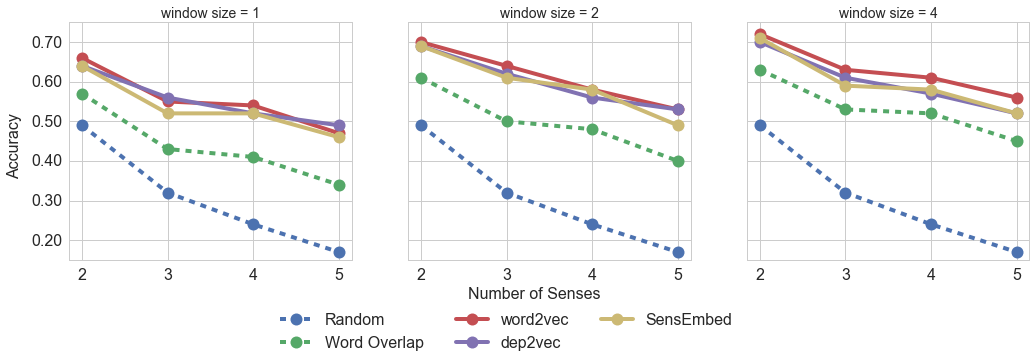}
\captionsetup{font=small}
\caption{Average performance across parts-of-speech per number of senses and context window.}
\label{results_linear_all_avg}
\end{figure*}

The perhaps most interesting observation in Table~\ref{results_linear_window_all} is that \texttt{word2vec} and \texttt{dep2vec} are performing as well or better than {\textsc{Sens}\textsc{Embed}} despite the fact that the former conflate the senses of a polysemous lexeme in a single vector representation. Figure~\ref{results_linear_all_avg} shows the average performance of all models across parts-of-speech per number of senses and for all context window sizes. 

\subsection*{{\textsc{Sens}\textsc{Embed}} and Babelfy}

One possible explanation for \Sensembed~not outperforming the other methods despite its cleaner encoding of different word senses in the above experiments is that at train time, it had access to sense labels from Babelfy. At test time on our task however, it did not have any sense labels available. We therefore sense tagged the 5-sense noun subtask with Babelfy and re-ran \Sensembed. As Table~\ref{results_sensembed_babelfy} shows, access to sense labels at test time did not give a substantive performance boost, representing further support for our hypothesis that composition in single-sense vector models might be sufficient to recover sense specific information.
\input{./_tables/babelfy.tex}

\subsection*{Frequency Range}
We chose the 2-sense noun subtask to estimate the degree sensitivity of target lexeme frequency on our task we merged the $[1, 1k)$ and $[1k, 10k)$, and $[50k, 100k)$ and $[100k, \infty )$ frequency bands from Figure~\ref{lexeme_frequencies}, and sampled an equal number of target words from each band. Table~\ref{results_frequency_range} reports the results for this experiment. All methods outperform the random and word overlap baseline and appear to be working better for less frequent lexemes. One possible explanation for this behaviour is that less frequent lexemes have fewer senses and potentially less subtle sense differences than more frequent lexemes, which would make them easier to discriminate by distributional semantic methods.
\input{./_tables/frequency_range.tex}

%% file: _tables/bow_1-4_no_adverbs.tex
\begin{table*}[!htb]
\centering
\small
\begin{tabular}{ l | c | c | c | c || c | c | c | c || c | c | c | c  }
\multicolumn{13}{c}{\textbf{Symmetric context window of size 1}}\\
& \multicolumn{4}{c||}{\textbf{Adjective}} & \multicolumn{4}{c||}{\textbf{Noun}} & \multicolumn{4}{c}{\textbf{Verb}}\\
\textbf{Senses} & 2 & 3 & 4 & 5 & 2 & 3 & 4 & 5 & 2 & 3 & 4 & 5  \\ \hline
\textbf{Random} & $0.53$ & $0.32$ & $0.25$ & $0.14$ & $0.47$ & $0.32$ & $0.23$ & $0.19$ & $0.47$ & $0.31$ & $0.23$ & $0.18$\\
\textbf{Word Overlap} & $0.63$ & $0.46$ & $0.47$ & $0.40$ & $0.55$ & $0.40$ & $0.37$ & $0.34$ & $0.54$ & $0.44$ & $0.38$ & $0.29$\\ \hline
\textbf{\texttt{word2vec}} & $\textbf{0.70}$ & $0.56$ & $\textbf{0.61}^{\dagger}$ & $0.54^{\dagger}$ & $\textbf{0.66}^{\ddagger}$ & $\textbf{0.52}^{\ddagger}$ & $\textbf{0.50}^{\ddagger}$ & $0.44^{\ddagger}$ & $\textbf{0.63}^{\ddagger}$ & $\textbf{0.56}^{\ddagger}$ & $\textbf{0.52}^{\ddagger}$ & $\textbf{0.43}^{\ddagger}$\\
\textbf{\texttt{dep2vec}} & $0.65$ & $\textbf{0.64}^{\ddagger}$ & $0.57$ & $\textbf{0.57}^{\ddagger}$ & $0.64^{\ddagger}$ & $0.50^{\ddagger}$ & $0.49^{\ddagger}$ & $\textbf{0.48}^{\ddagger}$ & $\textbf{0.63}^{\ddagger}$ & $0.55^{\ddagger}$ & $0.50^{\ddagger}$ & $\textbf{0.43}^{\ddagger}$\\
\textbf{{\textsc{Sens}\textsc{Embed}}} & $0.67$ & $0.54$ & $0.56$ & $0.56^{\dagger}$ & $0.64^{\ddagger}$ & $0.49^{\ddagger}$ & $\textbf{0.50}^{\ddagger}$ & $0.43^{\ddagger}$ & $0.62^{\dagger}$ & $0.53^{\ddagger}$ & $0.49^{\ddagger}$ & $0.38^{\dagger}$  \\ \hline\hline
\multicolumn{13}{c}{\textbf{Symmetric context window of size 2}}\\
& \multicolumn{4}{c||}{\textbf{Adjective}} & \multicolumn{4}{c||}{\textbf{Noun}} & \multicolumn{4}{c}{\textbf{Verb}}\\
\textbf{Senses} & 2 & 3 & 4 & 5 & 2 & 3 & 4 & 5 & 2 & 3 & 4 & 5  \\ \hline
\textbf{Random} & $0.53$ & $0.32$ & $0.25$ & $0.14$ & $0.47$ & $0.32$ & $0.23$ & $0.19$ & $0.47$ & $0.31$ & $0.23$ & $0.18$\\
\textbf{Word Overlap} & $0.66$ & $0.51$ & $0.55$ & $0.43$ & $0.59$ & $0.47$ & $0.43$ & $0.41$ & $0.58$ & $0.51$ & $0.45$ & $0.36$\\ \hline
\textbf{\texttt{word2vec}} & $0.70$ & $0.64^{\dagger}$ & $0.58$ & $0.55$ & $\textbf{0.71}^{\ddagger}$ & $\textbf{0.63}^{\ddagger}$ & $\textbf{0.59}^{\ddagger}$ & $0.54^{\ddagger}$ & $\textbf{0.68}^{\ddagger}$ & $0.64^{\ddagger}$ & $\textbf{0.58}^{\ddagger}$ & $\textbf{0.49}^{\ddagger}$\\
\textbf{\texttt{dep2vec}} & $0.71$ & $\textbf{0.65}^{\ddagger}$ & $0.58$ & $\textbf{0.57}^{\ddagger}$ & $0.70^{\ddagger}$ & $0.57^{\ddagger}$ & $0.55^{\ddagger}$ & $\textbf{0.55}^{\ddagger}$ & $0.66^{\dagger}$ & $0.64^{\ddagger}$ & $0.54^{\dagger}$ & $0.46^{\dagger}$\\
\textbf{{\textsc{Sens}\textsc{Embed}}} & $\textbf{0.72}^{\ddagger}$ & $0.62$ & $\textbf{0.61}$ & $0.52$ & $0.69^{\ddagger}$ & $0.56^{\dagger}$ & $0.57^{\ddagger}$ & $0.51^{\dagger}$ & $0.67^{\ddagger}$ & $\textbf{0.65}^{\dagger}$ & $0.57$ & $0.45$  \\\hline\hline
\multicolumn{13}{c}{\textbf{Symmetric context window of size 4}}\\
& \multicolumn{4}{c||}{\textbf{Adjective}} & \multicolumn{4}{c||}{\textbf{Noun}} & \multicolumn{4}{c}{\textbf{Verb}}\\
\textbf{Senses} & 2 & 3 & 4 & 5 & 2 & 3 & 4 & 5 & 2 & 3 & 4 & 5  \\ \hline
\textbf{Random} & $0.53$ & $0.32$ & $0.25$ & $0.14$ & $0.47$ & $0.32$ & $0.23$ & $0.19$ & $0.47$ & $0.31$ & $0.23$ & $0.18$\\
\textbf{Word Overlap} & $0.67$ & $0.55$ & $0.58$ & $0.51$ & $0.62$ & $0.50$ & $0.49$ & $0.45$ & $0.59$ & $0.55$ & $0.50$ & $0.40$\\ \hline
\textbf{\texttt{word2vec}} & $0.71$ & $0.65^{\dagger}$ & $\textbf{0.65}$ & $\textbf{0.57}$ & $\textbf{0.73}^{\ddagger}$ & $\textbf{0.61}^{\ddagger}$ & $\textbf{0.62}^{\ddagger}$ & $\textbf{0.57}^{\ddagger}$ & $\textbf{0.71}^{\ddagger}$ & $\textbf{0.62}^{\dagger}$ & $\textbf{0.57}$ & $\textbf{0.53}^{\ddagger}$\\
\textbf{\texttt{dep2vec}} & $0.72$ & $\textbf{0.66}^{\dagger}$ & $0.60$ & $0.54$ & $0.71^{\ddagger}$ & $0.55$ & $0.56^{\dagger}$ & $0.53^{\dagger}$ & $0.67$ & $0.62$ & $0.54$ & $0.50$\\
\textbf{{\textsc{Sens}\textsc{Embed}}} & $\textbf{0.75}$ & $0.59$ & $0.62$ & $0.55$ & $0.69^{\ddagger}$ & $0.57^{\dagger}$ & $0.58^{\ddagger}$ & $0.53^{\dagger}$ & $0.68^{\ddagger}$ & $0.62^{\dagger}$ & $0.55$ & $0.47$  \\
\end{tabular}
\captionsetup{singlelinecheck=false,font=small,labelsep=newline}
\caption{Performance overview for all parts-of-speech and number of senses, $\ddagger$ statistically significant at the $p<0.01$ level in comparison to the Word Overlap baseline; $\dagger$ statistically significant at the $p<0.05$ level in comparison to the Word Overlap baseline.}
\label{results_linear_window_all}
\end{table*} 

%% file: _tables/babelfy.tex
\begin{table}[!htb]
\centering
\small
\begin{tabular}{ l | c | c | c}
\multicolumn{4}{c}{\textbf{Noun - 5 Senses}}\\
\textbf{Context Window Size} & \textbf{1} & \textbf{2} & \textbf{4} \\\hline
\textbf{\texttt{word2vec}} & 0.44 & 0.54 & \textbf{0.57} \\
\textbf{\texttt{dep2vec}} & \textbf{0.48} & \textbf{0.55} & 0.53 \\
\textbf{\Sensembed} & 0.43 & 0.51 & 0.53 \\
\textbf{\Sensembed~\& Babelfy} & 0.45 & 0.49 & 0.54 \\
\end{tabular}
\captionsetup{singlelinecheck=false,font=small,labelsep=newline}
\caption{Results on the 5-sense noun subtask with \Sensembed~having access to Babelfy sense labels at test time.}
\label{results_sensembed_babelfy}
\end{table} 

%% file: _tables/frequency_range.tex
\begin{table}[!htb]
\centering
\small
\begin{tabular}{ l | c | c | c}
\multicolumn{4}{c}{\textbf{Noun - 2 Senses, context window size = 2}}\\
\textbf{Frequency Band} & \textbf{$< 10k$} & \textbf{$10k$ -- $50k$} & \textbf{$\geq 50k$} \\\hline
\textbf{Random} & 0.51 & 0.51 & 0.51 \\
\textbf{Word Overlap} & 0.66 & 0.60 & 0.56 \\\hline
\textbf{\texttt{word2vec}} & \textbf{0.81} & 0.64 & \textbf{0.66} \\
\textbf{\texttt{dep2vec}} & 0.77 & 0.67 & \textbf{0.66} \\
\textbf{\Sensembed} & 0.74 & \textbf{0.68} & 0.60 \\
\end{tabular}
\captionsetup{singlelinecheck=false,font=small,labelsep=newline}
\caption{Results on a subsample of the 2-sense noun subtask across frequency bands.}
\label{results_frequency_range}
\end{table} 

%% file: discussion.tex
Our results suggest that pointwise addition in a single-sense vector model such as \texttt{word2vec} is able to discriminate the sense of a polysemous lexeme in context in a surprisingly effective way and represents a strong baseline for future work.  Distributional composition can therefore be interpreted as a process of contextualising the meaning of a lexeme. This way, composition does not only act as a way to represent the meaning of a phrase as a whole, but also as a local discriminator for any lexemes in the phrase. For example the composed representation of \emph{dry clothes} should only keep contexts that \emph{dry} shares with \emph{clothes} while suppressing contexts it shares with \emph{weather} or \emph{wine}. Hence, one would expect the same to happen with a polysemous lexeme such as \emph{bank} in the context of \emph{river} and \emph{account}, respectively.

Recent work by Arora et al.~\shortcite{Arora_2016} has shown that the different senses of a polysemous lexeme reside in a linear substructure within a single vector and are recoverable by sparse coding. There is furthermore evidence that additive composition in low-dimensional word embeddings approximates an intersection of the contexts of two distributional word vectors~\cite{Tian_2015}. It therefore seems plausible that an intersective composition function should be able to recover sense specific information.

To qualitatively analyse this hypothesis we used the \texttt{word2vec} and \Sensembed~vectors to compose a small number of example phrases by pointwise addition and calculated their top 5 nearest neighbours in terms of cosine similarity. For \Sensembed~we manually sense tagged the phrases with the appropriate BabelNet sense labels prior to composition. We omitted the BabelNet sense labels in the neighbour list for brevity, however they were consistent with the intended sense in all cases. Table~\ref{composition_neighbours} supports the view of composition as a way of contextualising the meaning of a lexeme. 
\input{./_tables/composition_neighbours.tex}
In all cases in our example the \texttt{word2vec} neighbours reflect the intended sense of the polysemous lexeme, providing evidence for the linear substructure of word senses in a single vector as discovered by Arora et al.~\shortcite{Arora_2016}, and suggesting that distributional composition is able to recover sense specific information from a polysemous lexeme. The very fine-grained sense-level vector space of \Sensembed~is giving rise to a very focused neighbourhood, however there does not seem to be any advantage over \texttt{word2vec} from a qualitative point of view when using simple additive composition.

%

%% file: _tables/composition_neighbours.tex
\begin{table*}[!htb]
\centering
\small
\resizebox{\textwidth}{!}{
\begin{tabular}{ l | l | l}
\textbf{Phrase} & \textbf{\texttt{word2vec} neighbours} & \textbf{\Sensembed~neighbours} \\\hline
\emph{river bank} & bank, river, creek, lake, rivers & bank, river, stream, creek, river basin \\ \hline
\emph{bank account} & account, bank, accounts, banks, citibank & bank, banks, the bank, pko bank polski, handlowy\\\hline
\emph{dry weather} & weather, dry, wet weather, wet, unreasonably warm & dry, weather, humid, cold, cool \\\hline
\emph{dry clothes} & dry, clothes, clothing, rinse thoroughly, wet & dry, clothes, warm, cold, wet\\\hline
\emph{capital city} & capital, city, cities, downtown, town & city, capital, the capital city, town, provincial capital\\\hline
\emph{capital asset} & capital, asset, assets, investment, worth & capital, asset, investment, assets, investor\\\hline
\emph{power plant} & plant, power, plants, coalfired, megawatt & power, plant, near-limitless, pulse-power, power of the wind \\\hline
\emph{garden plant} & plant, garden, plants, gardens, vegetable garden & plant, garden, plants, oakville assembly, solanaceous\\\hline
\emph{window bar} & bar, window, windows, doorway, door & window, bar, windows, glass window, wall\\\hline
\emph{sandwich bar} & bar, sandwich, restaurant, burger, diner & sandwich, bar, restaurant, hot dog, cake \\\hline
\emph{gasoline tank} & gasoline, tank, fuel, gallon, tanks & gasoline, tank, fuel, petrol, kerosene\\\hline
\emph{armored tank} & armored, tank, tanks, M1A1 Abrams, armored vehicle & armored, armoured, tank, tanks, light tank\\\hline
\emph{desert rock} & rock, desert, rocks, desolate expanse, arid desert & desert rock, the desert, deserts, badlands\\\hline
\emph{rock band} & rock, band, rockers, bands, indie rock & band, rock, group, the band, rock group \\\hline
\end{tabular}}
\captionsetup{singlelinecheck=false,font=small,labelsep=newline}
\caption{Nearest neighbours of composed phrases for \texttt{word2vec} and \Sensembed. Distributional composition in \texttt{word2vec} is able to recover sense specific information remarkably well. Some neighbours are phrases because they have been encoded as a single token in the original vector space.}
\label{composition_neighbours}
\end{table*} 

%% file: relatedwork.tex
The perhaps most popular tasks for evaluating the ability of a model to capture or encode the different senses of a polysemous lexeme in a given context are the english lexical substitution task~\cite{McCarthy_2007} and the Microsoft sentence completion challenge~\cite{Zweig_2011}. Both tasks require any model to fill an appropriate word into a pre-defined slot in a given sentential context. The sentence completion challenge provides a list of candidate words while the english lexical substitution task does not. However, neither task focuses on polysemy and  the english lexical substitution task conflates the problems of discriminating word senses and finding meaning preserving substitutes. 

Dictionary definitions have previously been used to evaluate compositional distributional semantic models where the goal is to match a dictionary entry with its corresponding definition~\cite{Kartsaklis_2012,Polajnar_2014}. These datasets are commonly set up as retrieval tasks, but generally do not test the ability of a model to disambiguate a polysemous word in context, or discriminate multiple definitions of the same word. 

Our task also provides a novel evaluation for compositional distributional semantic models, where the predominant strategy is to estimate the similarity of two short phrases~\cite{Bernardi_2013,Grefenstette_2011b,Kartsaklis_2014,Mitchell_2008,Mitchell_2010} or sentences~\cite{Agirre_2016,Huang_2012,Marelli_2014} in comparison to human provided gold-standard judgements. One problem with these similarity tasks is that the similarity or relatedness of two sentences is very difficult to judge --- especially on a fine-grained scale --- even for humans. This frequently results in a relatively high variance of judgements and low inter-annotator agreement \cite{Batchkarov_2016}.  The short phrase datasets typically have a fixed structure that only test a very small fraction of the possible grammatical constructions in which a lexeme can occur, and furthermore provide very little context. The use of full sentences remedies the lack of context and grammatical variation, however can still contain a significant level of noise due to the automatic construction of the dataset or the variance in human ratings. In contrast, our task is not set up as a sentence similarity task and therefore avoids the use of human similarity judgements. 

Our task is similar to word-sense induction (WSI), however we only focus on discriminating the sense of a polysemous lexeme in context rather than inducing a set of senses from raw data and appropriately tagging subsequent occurrences of polysemous instances with the inferred inventory. Separating the sense discrimination task from the problem of sense induction has the advantage of making our task applicable to evaluating compositional distributional semantic models in order to test their ability to appropriately contextualise a polysemous lexeme. Due to not requiring any models to perform an extra step for sense induction, our task is easier to evaluate as no matching between sense clusters identified by a model and some gold standard sense classes needs to be performed, as typically proposed in the WSI literature~\cite{Agirre_2007,Manandhar_2010}. 

Most closely related to our task are the Stanford Contextual Word Similarity (SCWS) dataset by Huang et al.~\shortcite{Huang_2012} and the Usage Similarity (USim) task by Erk et al.~\shortcite{Erk_2009b}. The goal in both tasks is to estimate the similarity of two polysemous words in context in comparison to human provided gold standard judgements. In the SCWS dataset typically two different lexemes are considered whereas in USim and our task the same lexemes with different contexts are compared. Instead of relying on crowd-sourced human gold-standard similarity judgements, which can be prone to a considerable amount of noise\footnote{For example the average standard deviation of human ratings in the SCWS dataset is $\approx$$3$ on a 10-point scale, and can be  up to 4--5 in some cases.}, we leverage the high-quality content of available english dictionaries. Furthermore, our task is not formulated as estimating the similarity between two lexemes in context, but identifying the sentences that use the same sense of a given polysemous lexeme.

%% file: conclusion.tex
While elementary multi-sense representations of words might capture a more fine grained semantic picture of a polysemous word, that advantage does not appear to transfer to distributional composition in a straightforward way. Our experiments on a popular phrase similarity benchmark and our novel word-sense discrimination task have demonstrated that semantic composition does not appear to benefit from a fine grained sense inventory, but that the ability to contextualise a polysemous lexeme in single-sense vector models is sufficient for superior performance. We furthermore have provided qualitative and quantitative evidence that an intersective composition function such as pointwise addition for neural word embeddings is able to discriminate the meaning of a word in context, and is able to recover sense specific information remarkably well. 

Lastly, our experiments have uncovered an important question for multi-sense vector models, namely how to exploit the fine-grained sense level representations for distributional composition.  Our novel word-sense discrimination task provides an excellent testbed for compositional distributional semantic models, both following a single-sense or multi-sense vector modelling paradigm, due to its focus on assessing the ability of a model to appropriately contextualise the meaning of a word. Our task furthermore provides another evaluation option away from intrinsic evaluations which are based on often noisy human similarity judgements, while also not being embedded in a downstream task. 

In future work we aim to extend our evaluation to more complex compositional distributional semantic models such as the lexical function model~\cite{Paperno_2014} or the Anchored Packed Dependency Tree framework~\cite{Weir_2016}. We furthermore want to investigate how far the sense-discriminating ability of composition can be leveraged for other tasks.

%% file: acknowledgments.tex
We would like to thank our anonymous reviewers for their helpful comments.